\documentclass{article} % For LaTeX2e
\usepackage{iclr2027_conference,times}

\usepackage{amsmath,amsfonts,bm}

\def\eqref#1{equation~\ref{#1}}
\def\1{\bm{1}}

\DeclareMathAlphabet{\mathsfit}{\encodingdefault}{\sfdefault}{m}{sl}
\SetMathAlphabet{\mathsfit}{bold}{\encodingdefault}{\sfdefault}{bx}{n}

\usepackage{hyperref}
\usepackage{url}
\usepackage{xcolor}
\usepackage{booktabs}
\usepackage{makecell}
\usepackage{graphicx}
\usepackage{listings}  % required for \lstset and lstlisting
\usepackage{amsmath}
\usepackage{bbm}
\usepackage{wrapfig}
\usepackage{placeins}
\usepackage{multirow}
\newcommand{\na}{\textemdash}
\newcommand{\hs}[2]{#1/#2}
\newcommand{\staticonly}[1]{\na/#1}

\providecommand{\todo}[1]{}
\renewcommand{\todo}[1]{\textcolor{red}{\textbf{[TODO: #1]}}}

\definecolor{rankonefill}{HTML}{B7E4C7}
\definecolor{rankoneborder}{HTML}{2D6A4F}

\definecolor{ranktwofill}{HTML}{CFE8FF}
\definecolor{ranktwoborder}{HTML}{3A6EA5}

\definecolor{rankthreefill}{HTML}{FFE0B2}
\definecolor{rankthreeborder}{HTML}{C77800}

\newcommand{\rankone}[1]{%
  \begingroup
  \setlength{\fboxsep}{1.1pt}%
  \setlength{\fboxrule}{0.35pt}%
  \fcolorbox{rankoneborder}{rankonefill}{\strut\color{black}#1}%
  \endgroup
}

\newcommand{\ranktwo}[1]{%
  \begingroup
  \setlength{\fboxsep}{1.1pt}%
  \setlength{\fboxrule}{0.35pt}%
  \fcolorbox{ranktwoborder}{ranktwofill}{\strut\color{black}#1}%
  \endgroup
}

\newcommand{\rankthree}[1]{%
  \begingroup
  \setlength{\fboxsep}{1.1pt}%
  \setlength{\fboxrule}{0.35pt}%
  \fcolorbox{rankthreeborder}{rankthreefill}{\strut\color{black}#1}%
  \endgroup
}
\renewcommand{\rankone}[1]{#1}
\renewcommand{\ranktwo}[1]{#1}
\renewcommand{\rankthree}[1]{#1}

\title{ChronoGraph: Functional 4D Scene Graphs with Vision-Language Models for Interaction Understanding and Grounded Planning
}
\author{
{\bfseries
Chenyangguang Zhang$^{1}$\thanks{Equal contribution.}\quad
Malgorzata Gwiazda$^{1,2}$\footnotemark[1]\quad
Guanlong Jiao$^{3}$\quad
Yuanchen Ju$^{4}$} \\
{\bfseries
Federico Tombari$^{2,5}$\quad
Koushil Sreenath$^{4}$\quad
Marc Pollefeys$^{1,6}$\quad
Sunghwan Hong$^{1}$} \\[0.4em]
\normalfont
$^{1}$ETH Zurich \quad
$^{2}$Technical University of Munich \quad
$^{3}$University of British Columbia \\[-0.1em]
$^{4}$University of California, Berkeley \quad
$^{5}$Google \quad
$^{6}$Microsoft
}

\newif\ifarxiv
\arxivtrue       % Public arXiv version with authors
\ifarxiv
    \iclrfinalcopy % Makes the author list visible and removes review rulers
\fi

\begin{document}

\maketitle
\ifarxiv
    \lhead{Preprint. Under review.}
\fi

\vspace{-10mm}
\begin{figure}[htp]
    \centering
\includegraphics[width=\textwidth]{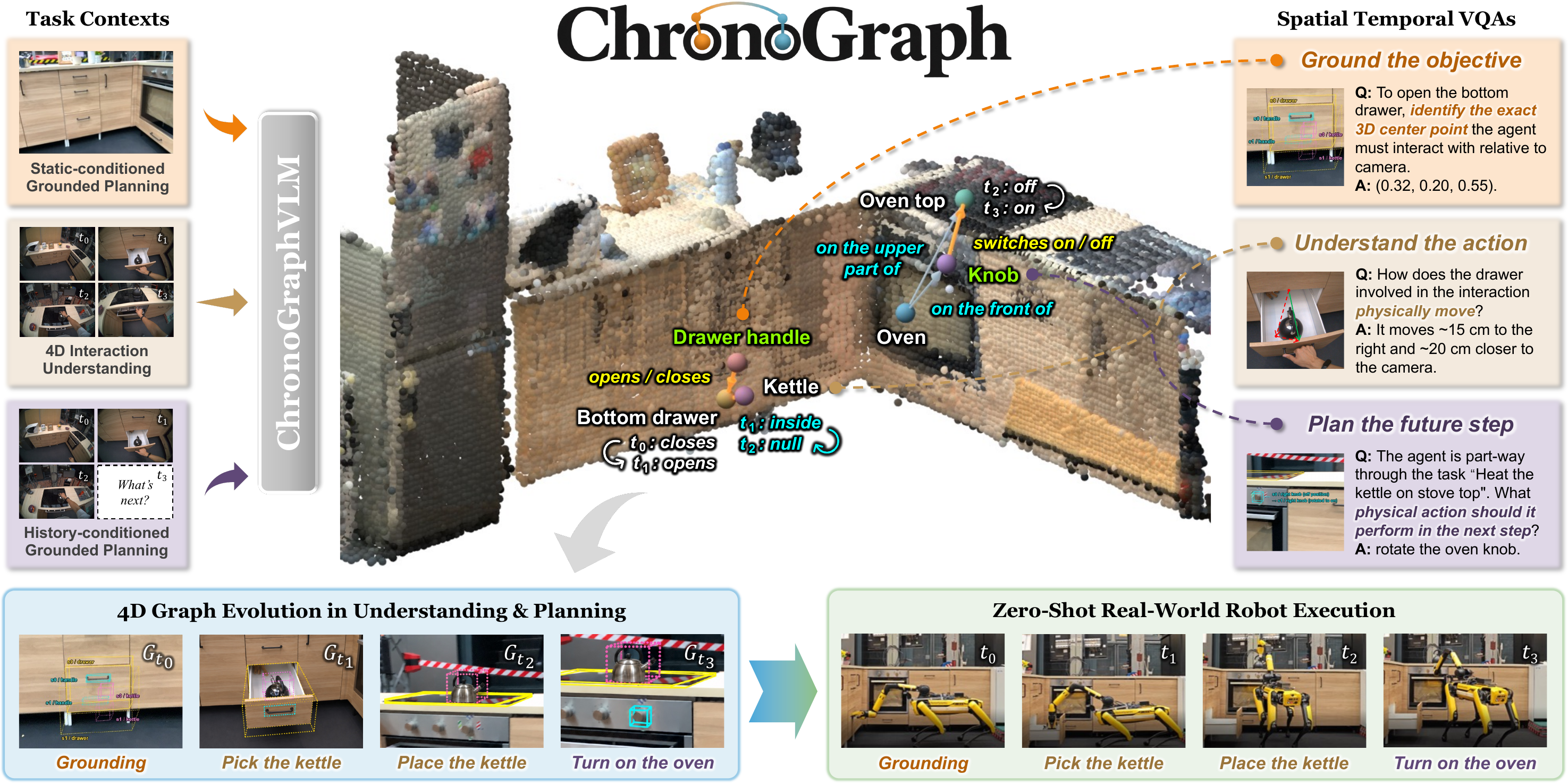}
\vspace{-5mm}
  \caption{\textbf{ChronoGraph connects 4D interaction understanding
with spatially grounded planning.}
ChronoGraphVLM generates functional 4D scene graphs linking
affordance-level actions to semantic and geometric state changes
for interaction-grounded VQA and planning.
Its predicted plans and affordance grounding guide zero-shot
real-world mobile manipulation.}
    \label{fig:teaser}
\end{figure}

\begin{abstract}
Embodied agents must determine where to act, anticipate the resulting scene changes, and interpret observed outcomes to guide subsequent actions.
This requires connecting \textit{4D interaction understanding}, which
explains how past actions changed the scene, with \textit{spatially
grounded planning}, which determines how and where to act toward a
goal and anticipates the resulting scene changes.
We introduce \textbf{ChronoGraph}, a functional 4D scene graph that links actions on affordance parts to semantic and geometric state changes.
By representing observed and anticipated transitions in the same form, it provides a shared basis for understanding and planning.
We construct \textbf{ChronoGraphBench} through an automatic data engine that converts human-interaction videos and simulated robot trajectories into graph-annotated questions for training and evaluating Vision-Language Models (VLMs) on both tasks.
Using these annotations, we train \textbf{ChronoGraphVLM} by adapting pretrained VLMs in two stages.
Graph-as-Chain-of-Thought supervised fine-tuning teaches the models to reconstruct observed transitions and predict future ones as graph traces before answering.
Subsequent joint 4D graph reinforcement learning directly rewards graph properties and answer correctness.
Experiments across model scales show improvements over the corresponding pretrained baselines and zero-shot transfer to VLM4D.
Real-world demonstrations further show that graph-based planning and affordance grounding support mobile manipulation through existing robot skills without additional fine-tuning.
\footnote{Code and data will be released.}
\end{abstract}

\section{Introduction}
\label{sec:intro}
Embodied tasks demand more than recognizing objects or describing their arrangement. 
When retrieving an item from a drawer, a robot must locate the handle and anticipate that pulling it will expose the contents. 
After acting, it must inspect and understand whether the drawer opened sufficiently and where the item can now be grasped. 
Its next decision depends on interpreting the actual outcome and identifying the affordances available in the updated scene.
4D understanding and grounded planning thus address complementary sides of the same interaction: one interprets realized semantic and geometric effects, while the other predicts the next action on a grounded affordance and anticipates future changes based on the interpreted current observation. 

Existing benchmarks for Vision-Language Models (VLMs) largely assess these capabilities separately.
Evaluations of 3D reasoning emphasize spatial attributes and relations~\citep{chen2024spatialvlm,cheng2024spatialrgpt,delitzas2024scenefun3d}, while 4D benchmarks examine how objects and scenes change over time~\citep{zhou2026learning,huang2026thinking,yin2026mllm}, focusing on observed object-level dynamics rather than interaction affordances. 
Planning benchmarks investigate action selection and task decomposition~\citep{ju2025momagraph,li2024muep,rana2023sayplan}, but rarely consider affordance-level grounding or successor-state prediction. 
This separation leaves the connection between understanding an interaction and planning the next one underexplored, particularly when an action changes the spatial configuration or accessibility of relevant affordances.

To make this connection explicit, we introduce \textbf{ChronoGraph}, a \emph{functional 4D scene graph} that provides a common representation for understanding and planning.
Object and affordance-part nodes encode semantic states and 3D attributes, while edges describe functional and spatial relations.
Action-linked transitions specify how manipulating a particular affordance changes the scene.
A graph sequence can therefore explain observed interactions or describe anticipated outcomes using the same entities, relations, and state transitions.
This shared structure provides a basis for learning both how interactions unfold and how their consequences inform subsequent actions.

Using this representation, we construct \textbf{ChronoGraphBench} from egocentric human-interaction videos~\citep{perrett2025hd,engelbracht2025hoi} and simulated robot trajectories~\citep{nasiriany2026robocasa365}. 
Our automatic data engine recognizes interactions in the recordings, generates graph annotations, and derives interaction-grounded Visual Question Answering (VQA) pairs. 
The benchmark comprises two tracks: \emph{4D interaction understanding}, which localizes observed interactions and recovers their actions, affordances, and semantic and geometric state changes, and \emph{spatially grounded planning}, which predicts goal-directed actions and their consequences. 
History-conditioned planning requires interpreting preceding interactions to determine the current state and subsequent steps~\citep{chen2023egoplan,sermanet2024robovqa}. 
Static-conditioned planning instead starts from a current image and additionally requires explicit grounding of execution locations.

We further turn these annotations into structured reasoning supervision for pretrained VLMs through a two-stage training pipeline, yielding \textbf{ChronoGraphVLM}.
We first develop Graph-as-Chain-of-Thought supervised fine-tuning, which teaches models to reconstruct observed transitions and predict future ones as graph traces before answering.
While this stage establishes the reasoning format, imitating a trace does not directly optimize its semantic and geometric correctness. 
We therefore follow it with joint 4D graph reinforcement learning, which provides feedback on graph semantic and geometric properties as well as answer correctness. 

Experiments on ChronoGraphBench demonstrate improvements over the corresponding pretrained baselines across model scales in both interaction understanding and planning, and these gains extend to zero-shot evaluation on VLM4D~\citep{zhou2025vlm4d}. 
Real-world mobile-manipulation demonstrations further show that the explicitly generated functional 4D scene graph, together with the learned planning and affordance-grounding capabilities, supports sequential execution through existing robot skills without any additional fine-tuning. In summary, our contributions are:
\begin{itemize}
    \item ChronoGraph, a functional 4D scene graph formulation that connects affordance-level interaction understanding with spatially grounded planning through observed and predicted action-conditioned state changes.
    \item ChronoGraphBench with a graph-based data engine, supporting training and evaluation of the interleaved understanding and planning tasks.
    \item ChronoGraphVLM from a training pipeline combining Graph-as-Chain-of-Thought supervised fine-tuning with joint 4D graph reinforcement learning, demonstrating improvements across model scales and benchmarks alongside transferring to robotic applications.
\end{itemize}
\section{Related Work}
\noindent\textbf{3D and 4D Understanding with VLMs.}
Early benchmarks evaluate spatial understanding through visual question answering~\citep{azuma2022scanqa,ma2022sqa3d,chen2024spatialvlm,cheng2024spatialrgpt}, while later efforts extend to multi-view and broader 3D reasoning~\citep{yang2025thinking,ma2026spatialreasoner,yang2026mmsi,jia2026omnispatial}. Image-based methods improve these capabilities through specialized training, preference optimization, and multi-image reasoning~\citep{zhang2026flatland,shen2026fine,xu2026multi,ouyang2025spacer}. Other approaches incorporate geometry through 3D scene inputs~\citep{huang2024chat,deng20253d}, depth cues~\citep{zhu2025llava,liu2026ssr}, or reconstruction priors~\citep{zheng2026learning,wu2026spatial,guo2025beyond}. Recent work addresses the temporal evolution of objects and scenes~\citep{zhou2025vlm4d,li2025sti,zhou2026learning,huang2026thinking,yin2026mllm}, alongside spatial traces and low-level embodied grounding~\citep{zhou2025robotracer,zhang2026embodied3dbench}. These directions primarily emphasize spatial relations, object and camera motion, or observed dynamics. We instead connect observed interactions to future actions at the level of functional affordance parts. Models must localize interactions, infer their semantic and spatial effects, and use the resulting state to ground the next action and predict its consequences.

\noindent\textbf{Embodied Planning with VLMs.}
VLM-based planning translates visual observations and language instructions into action sequences or spatial constraints~\citep{li2024muep,niu2024llarva,guo2024doremi,huang2023voxposer,huang2024rekep,zheng2025tracevla}. Graph-based approaches represent task-relevant entities and relations for planning and verification~\citep{rana2023sayplan,dai2024optimal,ekpo2024verigraph}, with MomaGraph~\citep{ju2025momagraph} incorporating objects, interactive parts, and their spatial and functional relations in form of plain text to describe static scenes. 
Our formulation connects this functional structure with observed and predicted interaction dynamics in temporal domain, linking affordance-level actions to semantic and geometric changes within a shared 4D grounded graph representation.
\section{Chronograph: Functional 4D Scene Graph for Understanding and Planning}
\label{sec:graph}

We extend functional 3D scene graphs~\citep{zhang2025open,hu2026hierarchical,fu2026funfact} to model the action-triggered 4D evolution of an interactive scene.
A functional 4D scene graph represents not only the objects and affordances present at a particular timestep, but also the semantic and spatial changes induced by interaction.
At time step $t$, we define the scene graph as:
\begin{equation}
G_t = (\mathcal{V}_t, \mathcal{E}_t), \qquad
\mathcal{V}_t = \mathcal{O}_t \cup \mathcal{P}_t,
\end{equation}
where $\mathcal{O}_t$ denotes task-relevant object nodes and $\mathcal{P}_t$ denotes affordance-part nodes.
Each node $v \in \mathcal{V}_t$ is associated with attributes of a name label, a semantic state, and 3D geometric information.
Each directed edge $e=(v_1,v_2,r) \in \mathcal{E}_t$ is labeled by a relation $r \in \mathcal{R}$, which contains both functional and spatial relations~\citep{ju2025momagraph}.
Thus, $G_t$ jointly represents what entities are present, which parts can be acted upon, and how their functional and spatial states are currently related.

An interaction $a_t$ is grounded on an affordance node $p_t$ and induces an action-conditioned transition
\begin{equation}
G_t \xrightarrow{a_t} G_{t+1},
\label{eq:graph_transition}
\end{equation}
where the relevant node states, geometric attributes, and relation edges have been updated.

For an entire task with goal $z$, the scene evolves through a sequence of atomic interactions:
\begin{equation}
\mathcal{G}_{0:T}
=
\left(
G_0 \xrightarrow{a_0} G_1
\xrightarrow{a_1} \cdots
\xrightarrow{a_{T-1}} G_T
\right).
\label{eq:long_horizon_transition}
\end{equation}
Each transition captures the local effect of one action, while the full chain represents the evolving functional and spatial state required to accomplish the goal.
Each action changes the objects, accessible affordances, and spatial configuration on which subsequent actions depend. We name this explicit 4D representation  \textbf{ChronoGraph}.

\section{ChronoGraphBench: Dataset and Benchmark}
\label{sec:data}

Connecting interaction understanding with planning requires annotations linking affordance-level actions to semantic and spatial changes. 
We utilize the representation of functional 4D scene graph $\mathcal{G}_{0:T}$ to construct the dataset and benchmark from egocentric human-interaction videos in HD-EPIC~\citep{perrett2025hd} and HOI!~\citep{engelbracht2025hoi}, alongside synthetic robot trajectories from RoboCasa~\citep{nasiriany2026robocasa365}.
These annotations support structured reasoning supervision and scalable generation of 4D-grounded VQA examples. 

As a result, our dataset comprises 1,669 interaction samples, 5,143 graph states, and 15,094 QA pairs from human egocentric recordings and robot simulations. 
It covers 357 object categories with questions spanning all conditioning settings and reasoning categories. 

\begin{figure}[t]
    \centering
    \includegraphics[width=\linewidth]{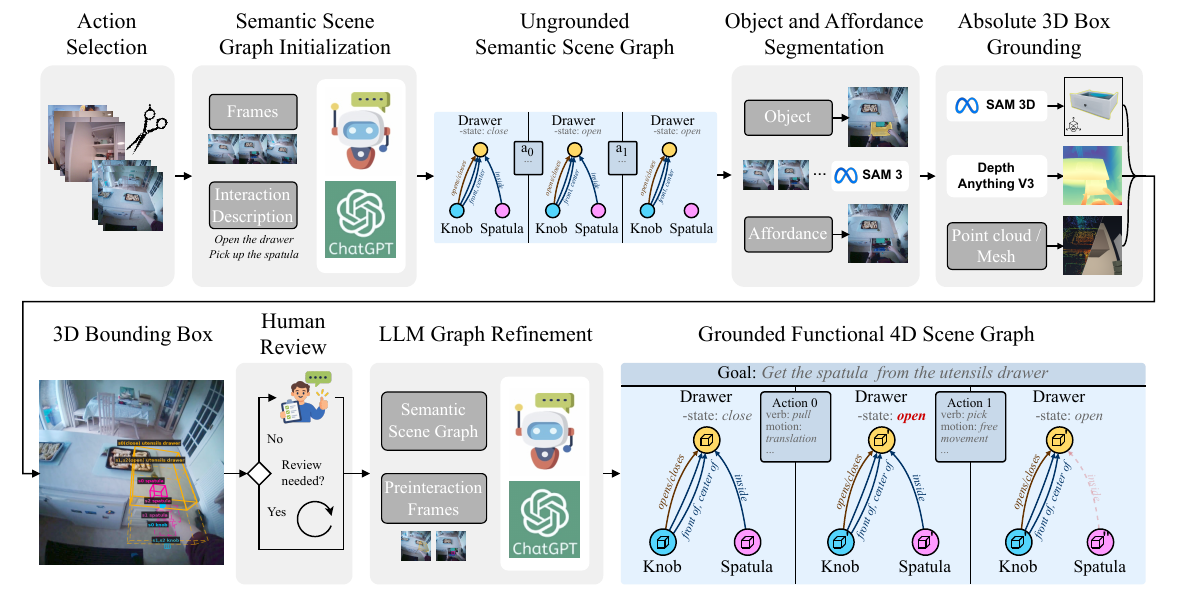}
    \vspace{-10mm}
    \caption{
\textbf{ChronoGraphBench annotation pipeline.}
We convert interaction recordings into functional 4D scene graphs linking affordance-level actions to semantic and geometric state changes.
The pipeline combines semantic graph initialization, object and affordance grounding, human verification, and graph refinement.
}
    \label{fig:automatic_4d_data_engine}
    \vspace{-6mm}
\end{figure}

\subsection{ChronoGraph Annotation}
\label{sec:data_engine}

All sources follow a shared annotation pipeline (Figure~\ref{fig:automatic_4d_data_engine}). The dataset metadata, annotations or simulator ground-truth are used to substitute corresponding pipeline components when available. 

\noindent\textbf{Semantic graph initialization.}
We first extract interaction clips with their temporal annotations, semantic labels, and descriptions. 
Given sampled videos and the metadata, GPT-5.5 is adopted to generate an initial semantic graph of entities, actions, semantic states, and relations. 
Few-shot examples specify the graph schema, establishing the semantic structure for subsequent grounding.

\noindent\textbf{Object and affordance grounding.}
For each interaction clip, SAM3~\citep{carion2025sam3segmentconcepts} segments the manipulated object and its affordance region. 
We reconstruct these regions with SAM-3D~\citep{sam3dteam2025sam3d3dfyimages} and estimate axis-aligned 3D bounding boxes for the corresponding graph nodes. 
To recover metric scale, we rescale each reconstruction using a low percentile of sparse metric point depths projected inside the mask, reducing contamination from background points. 
For masks without projected points, we use Depth Anything V3~\citep{lin2025depth} estimates calibrated to the metric point cloud via RANSAC. 
Boxes across all states share a world frame aligned with the scene's dominant horizontal direction.

\noindent\textbf{Human verification and graph refinement.}
Automatic checks trigger human review for incomplete or invalid graph fields, ambiguous visual or spatial evidence, and low-confidence vision model outputs.
We then refine the verified graphs with language descriptions.
GPT-5.5 generates referring expressions that distinguish interaction-relevant objects and, conditioned on the verified graph, describes executed actions, intended goals, and scene configurations.
Each completed record combines the semantic graph, object and affordance masks, 3D annotations, and descriptions for structured supervision and question generation.

\subsection{Question Construction and Evaluation Tasks}

\noindent\textbf{Question generation and validation.}
We generate graph-derived questions using 62 Python templates spanning atomic, short-horizon (2--4 sub-actions), and long-horizon (5+) interactions, given different characteristics of different time range.
Before full-dataset application, templates are iteratively refined on a representative subset across multiple review rounds to address phrasing, ambiguity, answerability, and distractor quality.
Questions must satisfy template-specific graph constraints.
Each multiple-choice question pairs a graph-derived answer with three semantically or functionally plausible distractors that are inconsistent with the annotations.
Distractor selection uses the dataset vocabulary, object statistics, and graph constraints, with GPT-5.6 Terra supplying additional candidates subject to answer-type, format, and mutual-exclusivity checks.
To reduce goal-based shortcuts in atomic action questions, we compare GPT-5.6 Terra's answer correctness with and without video frames under the same textual prompt.
Candidates are rejected if the frames provide no improvement across three repeated runs.
We do not apply this to longer-horizon questions, which inherently require integrating temporal evidence from the video, according to our development experiments.

\noindent\textbf{Evaluation settings.}
\emph{4D interaction understanding} evaluates realized transitions from videos, covering spatial tracking, action recognition, and functional and semantic state changes.
\emph{Static-conditioned planning} provides a pre-interaction image and goal for predicting the next action, target affordance, and successor state, including 2D/3D grounding of the interaction location.
\emph{History-conditioned planning} provides a goal and video prefix containing the first $k$ completed sub-actions, requiring inference of the resulting state and prediction of the next action and successor state.
This history provides observations of task-relevant objects and state changes that a single current egocentric frame may omit, \textit{e.g.}, after an agent retrieves food from a refrigerator and moves to a countertop, its current view may no longer reveal whether the refrigerator was left open.

\noindent\textbf{Reasoning categories.}
Questions fall into five categories. \emph{Action reasoning} covers observed or future actions, including sequences, next steps, and goal-relevant sub-actions. 
\emph{Affordance reasoning} identifies interactive object parts and their functions, while \emph{state change reasoning} concerns semantic transitions, affected entities, and transition times. 
\emph{Spatial reasoning} assesses object and affordance motion, including direction, relative configuration, changes in position, and camera-relative distance. 
Finally, \emph{2D/3D grounding} requires numerical predictions of target object or affordance centers in static-conditioned planning rather than multiple-choice answers.
\section{ChronoGraphVLM: Graph-Guided Learning for Understanding and Planning}

We train ChronoGraphVLM to generate a functional 4D scene graph trace before answering questions. 
Training comprises Graph-as-Chain-of-Thought supervised fine-tuning, which initializes structured graph generation and graph-conditioned answering, followed by joint 4D graph reinforcement learning, which optimizes graph consistency and answer quality. 

\subsection{Graph-as-Chain-of-Thought Supervised Fine-Tuning}

\noindent\textbf{Task-conditioned graph generation.} 
We represent each supervised example as an input--target pair $(x,y)$ and structure the target as an ordered \emph{graph--evidence--answer} sequence.
The graph first makes the task-relevant entities, affordances, actions, and state transitions explicit.
The evidence field then selects the graph properties that support the prediction, and the final answer is generated conditioned on this structured context. 
This ordering encourages the model to ground its answer in an explicit scene representation. 

Let $q$ denote the VQA question, $V_{\leq t}$ the available video, $I_t$ the current image, and $z$ the task goal. 
The model inputs and corresponding graph outputs for interaction understanding, history-conditioned planning, and static-conditioned planning are 
\begin{equation}
\resizebox{\columnwidth}{!}{$\displaystyle
x_{\mathrm{und}}=(V_{\leq t},q)
\longmapsto
(\hat{G}_{0:t},\hat{a}_{0:t-1}),
x_{\mathrm{hist}}=(V_{\leq t},z,q)
\longmapsto
(\hat{G}_{0:T},\hat{a}_{0:T-1}),
x_{\mathrm{stat}}=(I_t,z,q)
\longmapsto
(\hat{G}_{t:T},\hat{a}_{t:T-1})
$}
\label{eq:three_settings}
\end{equation}
where $T>t$ denotes the planning horizon. 
Interaction understanding reconstructs the observed graph transitions, whereas the two planning settings predict future actions and their resulting states, conditioned on the task goal. 
Each output trace follows the functional 4D scene graph representation $\mathcal{G}$ introduced in Section~\ref{sec:graph}. 
All graph attributes are serialized as text, while specifically 3D geometric information is serialized through a 3D center $\mathbf{p}_{i,s}\in\mathbb{R}^{3}$ for each node $i$ in state $s$. 
All centers share the coordinate system of the first input frame, or the current image for static-conditioned planning, which supports consistency of observed and predicted positions.

After generating the graph, the model outputs an evidence field $\hat{E}=[(\ell_j,c_j)]_{j=1}^{M}$, where $\ell_j$ identifies a graph property relevant to the question, such as a node name, semantic state or functional relation, \textit{etc.}, and $c_j$ is the content claimed for that property.
Ground-truth evidence is generated jointly with the VQA examples described in Section~\ref{sec:data}. 
Placing this field between the graph trace and the answer explicitly links the final prediction to its supporting graph content. 

\noindent\textbf{Training objective.}
Writing the serialized target as the token sequence $y=(y_1,\ldots,y_{|y|})$, we minimize the autoregressive negative log-likelihood:
\begin{equation} 
\mathcal{L}_{\mathrm{SFT}} = -\mathbb{E}_{(x,y)\sim\mathcal{D}_{\mathrm{SFT}}} \left[ \sum_{k=1}^{|y|} \log \pi_{\theta} \left( y_k \mid x,y_{<k} \right) \right], 
\label{eq:sft_objective} 
\end{equation} 
where $k$ indexes output tokens and $\pi_{\theta}$ denotes the VLM's conditional next-token distribution. 
The graph and evidence precede the answer, providing explicit context for answer generation.

\subsection{Joint 4D Graph Reinforcement Learning}

Token-level likelihood does not directly optimize structural consistency or task correctness.
We therefore refine the model using Group-Relative Policy Optimization~\citep{shao2024deepseekmath} with a composite reward that jointly evaluates graph properties and answer correctness:
\begin{equation}
R
=
\lambda_{\mathrm{fmt}}R_{\mathrm{fmt}}
+\lambda_{\mathrm{graph}}R_{\mathrm{graph}}
+\lambda_{\mathrm{mc}}R_{\mathrm{mc}}
+\lambda_{\mathrm{gnd}}R_{\mathrm{gnd}},
\label{eq:composite_reward}
\end{equation}
where the coefficients weight format validity, graph quality, and answer correctness.

\noindent\textbf{Prediction alignment.}
Inserted or omitted graph states can misalign all subsequent index-wise comparisons. 
We therefore apply Needleman--Wunsch alignment~\citep{needleman1970general} using graph-state similarity:
\begin{equation}
\operatorname{sim}(n_p,n_g)
=
\eta\,\mathcal{M}
\left(d_{p}^{\mathrm{state}},d_{g}^{\mathrm{state}}\right)
+
(1-\eta)\operatorname{softF1}(\mathcal{N}_p,\mathcal{N}_g),
\label{eq:state_alignment}
\end{equation}
where $p,g$ indicate the predicted and reference node, $\mathcal{M}$ measures semantic state agreement, $\mathcal{N}_p$ and $\mathcal{N}_g$ are the respective node name sets, and $\eta$ balances the terms. 
Soft F1 is the harmonic mean of soft precision and recall, each obtained by averaging the best cross-set name agreement in its respective direction. 
Semantic state descriptions distinguish consecutive states that share the same entities, while the monotone alignment preserves temporal order.

\noindent\textbf{Format and graph rewards.}
The format reward $R_{\mathrm{fmt}}$ checks the output wrapper, nonempty state sequence, contiguous indices, and graph state--action interleaving. 
The graph reward $R_{\mathrm{graph}}$ evaluates node names and edge types using soft F1, semantic state agreement between name-matched nodes, and graph state count. 
It also scores semantic and relational changes across aligned transitions. 
Action sequence which connects consecutive graph states is scored with equal credit to the correct affordance verb and target affordance node name.
For each name-matched node, the reward on each node's 3D center $\mathbf{p}_{i,s}$ is defined as $\exp\left[
-\frac{1}{2}
\left(\frac{d_{uv}}{\sigma_{uv}}\right)^2
-\frac{1}{2}
\left(\frac{d_z}{\sigma_z}\right)^2
\right],$
where $d_{uv}$ is the Euclidean error in normalized image coordinates, $d_z$ is the absolute depth error, and $\sigma_{uv}$ and $\sigma_z$ control the respective tolerances. 
To capture temporal motion beyond absolute position, we additionally compute a reward of 
$
\max\left(
0,
\frac{
\Delta\mathbf{c}_{\mathrm{gt}}^\top
\Delta\mathbf{c}_{\mathrm{pred}}
}{
\|\Delta\mathbf{c}_{\mathrm{gt}}\|_2
\|\Delta\mathbf{c}_{\mathrm{pred}}\|_2
}
\right),
$
where $\Delta\mathbf{c}=\mathbf{c}_{s+1}-\mathbf{c}_{s}$ denotes the center motion across corresponding transitions. 
Transitions with negligible reference motion are excluded because their direction is unstable.

\noindent\textbf{Answer rewards.}
Multiple-choice questions use exact-match accuracy to form $R_{\mathrm{mc}}$. 
Grounding questions use $R_{\mathrm{gnd}}=\exp(-\alpha d)$, where $d$ is the distance from the predicted point to the target 2D or 3D box, defined as zero inside it. 
This assigns full credit within the target and smoothly decreasing credit outside it.
\section{Experiments}

\subsection{Experimental Setup}

\noindent\textbf{Datasets and metrics.}
We partition the benchmark data into mutually exclusive subsets of the level of the household or simulation set-up for supervised fine-tuning, reinforcement learning and holdout evaluation, comprising $20\%$, $68\%$, and $12\%$ of the data, respectively.  
This deliberately short supervised fine-tuning stage primarily teaches the model the structured graph evidence output format and provides a stable initialization for reinforcement learning. 
It also reduces overfitting to the supervised trajectories and better preserves the general visual and linguistic capabilities of the pretrained model~\citep{chu2025sft}. 
The evaluation set contains 202 interaction samples, including 129 human and 73 robotic interactions, from which we construct 1,811 VQAs, with careful human verification.
The benchmark contains single-answer multiple-choice questions and open-numerical grounding questions. 
We evaluate multiple-choice questions using exact-match accuracy. 
For 2D and 3D grounding, we use the soft score $\exp(-\alpha d)$~\citep{zhang2026embodied3dbench}, where $d$ denotes the distance from the predicted point to the target box and is defined as zero when the prediction lies inside the box. 
We set $\alpha=8$ for normalized 2D coordinates and $\alpha=5$ for metric 3D coordinates.

\noindent\textbf{Baselines.}
We compare our method against three groups of baselines. 
First, we evaluate state-of-the-art commercial models, including GPT 6 Astra, 5.6 Sol, 5.6 Terra, Gemini 3.1 Pro, 3.8 Flash, Claude Opus 5, and Kimi K3, with the medium reasoning effort configuration. 
Second, we evaluate open-source VLMs from the Qwen model families across multiple parameter scales. 
Finally, we compare with specialized methods for 4D interaction understanding and embodied planning, including MomaGraph~\citep{ju2025momagraph}, SpatialVLM~\citep{chen2024spatialvlm}, DSR~\citep{zhou2026learning}, and MLLM-4D~\citep{yin2026mllm}.
All models are evaluated using the same input protocol: 
for 4D interaction understanding and history-conditioned planning, videos are uniformly sampled at 4 FPS;
for static-conditioned planning, each model receives the corresponding pre-interaction image.

\subsection{Main Results}

\begin{table*}[t]
\centering
\caption{\textbf{Main comparison on ChronoGraphBench for 4D interaction understanding and spatially grounded planning.}
Planning scores are reported as \emph{history-conditioned/static-conditioned}, with 2D/3D grounding evaluated only in the static-conditioned setting.
The overall average is the unweighted mean of per-question scores across all settings.}
\label{tab:main_results}

\setlength{\tabcolsep}{2.7pt}
\renewcommand{\arraystretch}{1.25}

\resizebox{\textwidth}{!}{%
\begin{tabular}{lcccccccccccc}
\toprule
& \multicolumn{5}{c}{\textbf{4D Interaction Understanding}}
& \multicolumn{6}{c}{\textbf{Spatially Grounded Planning}}
& \multicolumn{1}{c}{} \\
\cmidrule(lr){2-6}
\cmidrule(lr){7-12}

\textbf{Model}
& \makecell{Action\\Recognition}
& \makecell{Affordance\\Recognition}
& \makecell{Spatial\\Reasoning}
& \makecell{State Change\\Recognition}
& \textbf{Avg.}
& \makecell{Action\\Prediction}
& \makecell{Affordance\\Prediction}
& \makecell{Spatial\\Prediction}
& \makecell{State Change\\Prediction}
& \makecell{2D/3D\\Grounding}
& \textbf{Avg.}
& \makecell{\textbf{Overall}\\\textbf{Avg.}} \\
\midrule

\multicolumn{13}{l}{\textit{\textbf{Commercial Models}}} \\

GPT 6 Astra
& \rankone{93.8}
& \ranktwo{72.6}
& \rankone{75.5}
& \rankone{88.1}
& \rankone{85.5}
& \hs{\rankone{85.4}}{\rankthree{85.4}}
& \hs{\rankone{81.1}}{81.7}
& \hs{\rankone{78.1}}{\rankone{69.2}}
& \hs{\rankthree{85.3}}{\rankthree{96.6}}
& \staticonly{\rankone{67.9}}
& \hs{\rankone{81.3}}{\rankone{74.2}}
& \rankone{79.4} \\

GPT 5.6 Sol
& \rankthree{86.8}
& \ranktwo{72.6}
& \rankthree{69.6}
& \rankthree{83.5}
& \rankthree{80.1}
& \hs{\ranktwo{78.0}}{\rankthree{85.4}}
& \hs{\ranktwo{78.4}}{78.5}
& \hs{\ranktwo{75.0}}{57.8}
& \hs{\ranktwo{88.2}}{93.1}
& \staticonly{61.2}
& \hs{\ranktwo{78.4}}{68.1}
& 73.9 \\

GPT 5.6 Terra
& 84.9
& \rankone{74.2}
& 64.1
& 77.8
& 76.5
& \hs{\rankthree{75.6}}{82.6}
& \hs{67.6}{\ranktwo{83.9}}
& \hs{65.6}{54.0}
& \hs{61.8}{93.1}
& \staticonly{\rankthree{61.5}}
& \hs{67.3}{67.4}
& 70.9 \\

Gemini 3.1 Pro
& 85.3
& \rankone{74.2}
& 65.2
& 82.5
& 78.2
& \hs{\rankthree{75.6}}{\ranktwo{86.1}}
& \hs{\rankthree{73.0}}{\rankone{84.9}}
& \hs{\rankthree{67.7}}{\rankthree{63.5}}
& \hs{\ranktwo{88.2}}{\rankone{100.0}}
& \staticonly{61.0}
& \hs{73.6}{\rankthree{70.5}}
& \rankthree{73.8} \\

Gemini 3.8 Flash
& \ranktwo{91.9}
& \rankone{74.2}
& \ranktwo{72.8}
& \ranktwo{85.1}
& \ranktwo{83.4}
& \hs{\ranktwo{78.0}}{\rankone{86.8}}
& \hs{\ranktwo{75.7}}{\rankone{84.9}}
& \hs{\rankthree{67.7}}{\ranktwo{67.3}}
& \hs{\rankone{91.2}}{\ranktwo{98.3}}
& \staticonly{\ranktwo{66.2}}
& \hs{\rankthree{75.0}}{\ranktwo{73.7}}
& \ranktwo{77.6} \\

Claude Opus 5
& 70.9
& \rankthree{56.5}
& 54.3
& 70.1
& 65.0
& \hs{68.3}{80.6}
& \hs{62.2}{\rankthree{82.8}}
& \hs{61.5}{57.3}
& \hs{67.6}{93.1}
& \staticonly{51.7}
& \hs{63.9}{63.4}
& 64.1 \\
Kimi K3 2.8T
& \rankone{83.3}
& \rankthree{71.0}
& \rankone{64.1}
& \rankone{83.0}
& \rankone{77.1}
& \hs{\rankone{68.3}}{\ranktwo{86.1}}
& \hs{\rankone{73.0}}{78.5}
& \hs{\ranktwo{68.8}}{\ranktwo{58.8}}
& \hs{\rankone{82.4}}{\rankthree{93.1}}
& \staticonly{\rankthree{61.0}}
& \hs{\rankone{71.6}}{\rankthree{68.3}}
& \rankthree{72.1} \\
\midrule

\multicolumn{13}{l}{\textit{\textbf{Open-Source Models}}} \\

% Qwen-3.8 Flash Next 125B
% & 75.7
% & 65.0
% & 60.3
% & 75.7
% & 71.6
% & \hs{\ranktwo{59.1}}{82.9}
% & \hs{42.1}{77.8}
% & \hs{51.2}{\rankthree{52.6}}
% & \hs{33.3}{89.1}
% & \staticonly{45.0}
% & \hs{48.0}{59.6}
% & 63.7 \\

Qwen-3.8 27B
& 72.1
& 69.4
& 45.7
& 71.1
& 64.6
& \hs{53.7}{78.5}
& \hs{45.9}{81.7}
& \hs{56.2}{47.9}
& \hs{58.8}{91.4}
& \staticonly{45.9}
& \hs{54.3}{58.2}
& 60.2 \\

Qwen-3.5 27B
& 72.1
& \ranktwo{72.6}
& 54.9
& 72.2
& 67.6
& \hs{\rankthree{58.5}}{77.1}
& \hs{\rankthree{54.1}}{84.9}
& \hs{\rankthree{63.5}}{45.0}
& \hs{70.6}{\rankthree{93.1}}
& \staticonly{40.0}
& \hs{62.0}{55.1}
& 60.7 \\

\midrule

\multicolumn{13}{l}{\textit{\textbf{4D / Planning Experts}}} \\

MomaGraph 7B
& 55.8
& 61.3
& 43.5
& 64.9
& 55.6
& \hs{39.0}{77.8}
& \hs{27.0}{76.3}
& \hs{44.8}{36.0}
& \hs{\rankthree{73.5}}{87.9}
& \staticonly{33.8}
& \hs{45.2}{49.2}
& 51.2 \\

SpatialVLM 13B
& 43.8
& 40.3
& 32.1
& 34.0
& 37.7
& \hs{41.5}{56.9}
& \hs{32.4}{72.0}
& \hs{46.9}{28.9}
& \hs{29.4}{84.5}
& \staticonly{30.6}
& \hs{40.4}{42.1}
& 40.2 \\

DSR 7B
& 54.3
& 58.1
& 45.7
& 62.9
& 54.7
& \hs{41.5}{76.4}
& \hs{27.0}{78.5}
& \hs{46.9}{33.2}
& \hs{\ranktwo{79.4}}{82.8}
& \staticonly{31.8}
& \hs{47.6}{47.3}
& 50.2 \\

MLLM-4D 8B
& 58.1
& 66.1
& 41.8
& 63.9
& 56.2
& \hs{51.2}{74.3}
& \hs{\ranktwo{54.1}}{\rankthree{86.0}}
& \hs{43.8}{46.9}
& \hs{41.2}{86.2}
& \staticonly{33.5}
& \hs{46.6}{51.9}
& 53.0 \\
\midrule

\multicolumn{13}{l}{\textit{\textbf{Baselines v.s. Ours}}} \\

Qwen-3.5 9B
& 62.8
& 59.7
& 38.0
& 60.3
& 55.3
& \hs{48.8}{71.5}
& \hs{45.9}{79.6}
& \hs{55.2}{41.7}
& \hs{50.0}{89.7}
& \staticonly{29.1}
& \hs{51.4}{47.9}
& 51.2 \\

\textbf{ChronoGraphVLM 9B}
& \rankthree{78.3}
& \rankone{87.1}
& \ranktwo{62.5}
& \ranktwo{82.0}
& \ranktwo{75.9}
& \hs{53.7}{\rankone{88.9}}
& \hs{51.4}{\rankone{92.5}}
& \hs{\rankone{74.0}}{\rankone{60.7}}
& \hs{\ranktwo{79.4}}{\ranktwo{96.6}}
& \staticonly{\ranktwo{73.7}}
& \hs{\ranktwo{66.8}}{\rankone{76.5}}
& \rankone{75.2} \\

Qwen-3.5 4B
& 53.1
& 56.5
& 37.0
& 53.6
& 49.3
& \hs{46.3}{74.3}
& \hs{24.3}{82.8}
& \hs{46.9}{36.0}
& \hs{35.3}{86.2}
& \staticonly{36.8}
& \hs{40.9}{50.5}
& 48.9 \\

\textbf{ChronoGraphVLM 4B}
& \ranktwo{79.1}
& \rankone{87.1}
& \rankthree{63.0}
& \rankthree{76.3}
& \rankthree{74.8}
& \hs{51.2}{\rankthree{84.7}}
& \hs{51.4}{\ranktwo{87.1}}
& \hs{\rankone{74.0}}{49.3}
& \hs{\ranktwo{79.4}}{\rankone{98.3}}
& \staticonly{\rankone{74.9}}
& \hs{\rankthree{66.3}}{\ranktwo{73.2}}
& \ranktwo{73.0} \\
\bottomrule
\end{tabular}%
}
\vspace{-5mm}
\end{table*}

\begin{wraptable}{r}{0.3\columnwidth}
% \vspace{-20pt}
\centering
\caption{\textbf{Zero-shot transfer to VLM4D.}
We report multiple-choice accuracy on the egocentric split.}
\label{tab:vlm4d_zero_shot}
\small
\begin{tabular}{@{}l@{\hspace{35pt}}c@{}}
\toprule
\textbf{Method}
& \textbf{Acc. (\%)} $\uparrow$ \\
\midrule
Qwen3.5-9B
& 60.5 \\
Qwen3.5-4B
& 53.7 \\
\textbf{Ours-9B}
& 67.1 \\
\textbf{Ours-4B}
& 61.0 \\
\bottomrule
\end{tabular}
% \vspace{-10pt}
\end{wraptable}

Table~\ref{tab:main_results} compares 4D interaction understanding and both planning settings.
GPT-6 Astra leads the commercial baselines overall, while the Qwen series exhibits an overall scaling trend.
However, strong semantic predictions do not imply equally accurate spatial reasoning: in static-conditioned planning, GPT-6 Astra scores $96.6$ on state-change prediction but $69.2$ on spatial prediction.
Specialized 4D understanding or planning methods also remain limited under the unified evaluation, reaching at most $53.0$ overall.
ChronoGraphVLM improves understanding and both planning settings at both model scales, increasing the overall score by $24.0$ at 9B and $24.1$ at 4B over the corresponding baselines.
The resulting scores of $75.2$ and $73.0$ surpass all evaluated open-source and specialized baselines.
The category-level strengths align with the explicit semantic and geometric guidance from the graph.
Nevertheless, a substantial gap remains in history-conditioned action and affordance prediction.
These results distinguish improved interaction grounding from the still-challenging problem of selecting subsequent actions from an evolving interaction history.

\subsection{Zero-Shot Cross-Benchmark Results}

We evaluate zero-shot transfer on the egocentric split of VLM4D~\citep{zhou2025vlm4d} without additional fine-tuning.
As shown in Table~\ref{tab:vlm4d_zero_shot}, ChronoGraphVLM achieves $67.1\%$ accuracy at 9B and $61.0\%$ at 4B, improving over the corresponding pretrained initializations by $6.6\%$ and $7.3\%$.
Since VLM4D differs in both evaluation data and question construction, these gains suggest that graph-guided training develops transferable spatiotemporal reasoning rather than only improving performance on ChronoGraphBench-specific questions.
As shown in Section~\ref{sec:ablation}, this transfer relies on graph supervision, since answer-only fine-tuning reduces VLM4D accuracy below the pretrained model.

% As shown in Table~\ref{tab:vlm4d_zero_shot}, our method consistently improves zero-shot 4D understanding at both model scales. 
% Ours-9B outperforms its Qwen3.5-9B initialization by $6.6\%$, while Ours-4B yields a larger absolute gain of $7.3\%$. 
% Supervision through functional 4D scene graphs encourages the model to represent temporally evolving object states, interaction-relevant entities, and spatial changes explicitly. 
% These capabilities transfer to VLM4D even though its questions and evaluation data differ from those used during training. 
% The consistent improvements at both 4B and 9B scales further suggest that the gains arise from the proposed representation and training objectives rather than model capacity alone.

\subsection{Ablation Studies}\label{sec:ablation}
Table~\ref{tab:ablation} reports performance on ChronoGraphBench and zero-shot transfer to the egocentric split of VLM4D for the ablations of graph-guided reasoning and training with Qwen3.5-9B.

\begin{table*}[t]
\centering
\caption{
\textbf{Ablations of graph-guided reasoning and training with Qwen3.5-9B.}
ChronoGraphBench planning scores are reported as \emph{history-conditioned / static-conditioned}.
The final column reports zero-shot accuracy on the egocentric split of VLM4D.
}
\label{tab:ablation}

\setlength{\tabcolsep}{2.7pt}
\renewcommand{\arraystretch}{1.25}

\resizebox{\textwidth}{!}{%
\begin{tabular}{lccccccccccccc}
\toprule
& \multicolumn{5}{c}{\textbf{4D Interaction Understanding}}
& \multicolumn{6}{c}{\textbf{Spatially Grounded Planning}}
& \multicolumn{1}{c}{}
& \multicolumn{1}{c}{\textbf{VLM4D}} \\
\cmidrule(lr){2-6}
\cmidrule(lr){7-12}
\cmidrule(lr){14-14}

\textbf{Method}
& \makecell{Action\\Recognition}
& \makecell{Affordance\\Recognition}
& \makecell{Spatial\\Reasoning}
& \makecell{State Change\\Recognition}
& \textbf{Avg.}
& \makecell{Action\\Prediction}
& \makecell{Affordance\\Prediction}
& \makecell{Spatial\\Prediction}
& \makecell{State Change\\Prediction}
& \makecell{2D/3D\\Grounding}
& \textbf{Avg.}
& \makecell{\textbf{Overall}\\\textbf{Avg.}}
& \makecell{\textbf{Acc.}\\\textbf{(\%)}} \\
\midrule

\multicolumn{14}{l}{
\textit{\textbf{(a) Graph-Guided Reasoning}}
} \\

Qwen3.5-9B (Direct)
& 62.8 & 59.7 & 38.0 & 60.3
& 55.3
& \hs{48.8}{71.5}
& \hs{45.9}{79.6}
& \hs{55.2}{41.7}
& \hs{50.0}{89.7}
& \staticonly{29.1}
& \hs{51.4}{47.9}
& 51.2
& 60.5 \\

Qwen3.5-9B + Naive Thinking
& 68.3 & 65.6 & 59.0 & 59.7
& 63.8
& \hs{46.2}{76.9}
& \hs{44.4}{77.2}
& \hs{45.5}{47.6}
& \hs{76.4}{92.7}
& \staticonly{26.0}
& \hs{49.5}{50.6}
& 55.8
& 62.2 \\

Qwen3.5-9B + Graph-as-CoT
& 67.8 & 67.7 & 50.0 & 70.1
& 63.8
& \hs{51.2}{80.6}
& \hs{45.9}{77.4}
& \hs{58.3}{49.8}
& \hs{50.0}{87.9}
& \staticonly{42.4}
& \hs{53.3}{56.7}
& 59.0
& 63.9 \\
\midrule

\multicolumn{14}{l}{
\textit{\textbf{(b) Supervised Fine-Tuning}}
} \\

SFT: Answer Only
& 75.2 & 79.0 & 51.6 & 75.3
& 69.3
& \hs{56.0}{82.6}
& \hs{45.9}{81.7}
& \hs{59.4}{49.8}
& \hs{67.6}{93.1}
& \staticonly{64.3}
& \hs{57.7}{67.5}
& 67.0
& 49.4 \\

\textbf{Ours: Graph-as-CoT SFT}
& 76.0 & 82.3 & 55.4 & 71.1
& 69.8
& \hs{51.2}{86.1}
& \hs{48.6}{89.2}
& \hs{64.6}{55.5}
& \hs{79.4}{91.4}
& \staticonly{68.9}
& \hs{61.5}{72.0}
& 69.9
& 64.2 \\
\midrule

\multicolumn{14}{l}{
\textit{\textbf{(c) Reinforcement Learning}}
} \\

RL: Answer Reward Only
& 78.3 & 85.5 & 62.5 & 78.4
& 74.9
& \hs{56.0}{88.9}
& \hs{43.2}{89.2}
& \hs{70.8}{54.0}
& \hs{76.5}{94.8}
& \staticonly{72.2}
& \hs{63.9}{73.8}
& 73.0
& 65.4 \\

\textbf{Ours: Answer + Graph Reward}
& 78.3 & 87.1 & 62.5 & 82.0
& 75.9
& \hs{53.7}{88.9}
& \hs{51.4}{92.5}
& \hs{74.0}{60.7}
& \hs{79.4}{96.6}
& \staticonly{73.7}
& \hs{66.8}{76.5}
& 75.2
& 67.1 \\
\bottomrule
\end{tabular}%
}
\vspace{-5mm}
\end{table*}

\noindent\textbf{Graph-guided reasoning.}
Graph-as-CoT prompting instructs the pretrained model to identify task-relevant nodes and relations before answering, without requiring the complete functional 4D graph schema or metric 3D predictions. 
This training-free intervention improves the overall ChronoGraphBench score by $7.8$ over direct answering and by $3.2$ over the model's native free-form thinking mode. 
It also improves zero-shot VLM4D accuracy by $3.4\%$ and $1.7\%$ respectively. 
Notably, graph-guided models, including ours, do not produce longer outputs on average than models using free-form reasoning.
The improvement over free-form thinking indicates that explicitly organizing entities and relations produces more intelligent behavior even without parameter updates.

\noindent\textbf{Graph-as-chain-of-thought supervised fine-tuning.}
Relative to the baseline, answer-only supervision raises the ChronoGraphBench overall average from $51.2$ to $67.0$, but reduces zero-shot VLM4D accuracy from $60.5\%$ to $49.4\%$.
Graph-as-CoT instead improves both, reaching $69.9$ and $64.2\%$ respectively.
This contrast suggests that supervising intermediate entities and state transitions improves both performance on in-domain data and cross-benchmark generalization.

\noindent\textbf{Joint 4D graph reinforcement learning.}
Answer-only reward only increases the ChronoGraphBench overall score from $69.9$ to $73.0$ and VLM4D accuracy from $64.2\%$ to $65.4\%$, while including graph rewards raises them further to $75.2$ and $67.1\%$.
These results suggest that feedback on graph properties improving understanding and planning with cross-benchmark generalization.

\subsection{Application: Mobile Manipulation}
\label{sec:robot_experiments}

We deploy ChronoGraphVLM zero-shot on a Boston Dynamics Spot quadruped without environment- or task-specific fine-tuning.
Given an initial RGB observation and a high-level goal, the model generates a subaction sequence and a task-relevant functional scene graph.
Before each subaction, it uses the observation history to predict subsequent actions and grounds the target affordance in the current image.
The predicted location is converted to a 3D target using measured depth and registered in the robot's local map for execution through the corresponding motion primitive.
The robot then observes the updated scene and repeats this perception--grounding--action loop.

\begin{figure*}[t]
    \centering
    \includegraphics[width=\textwidth]{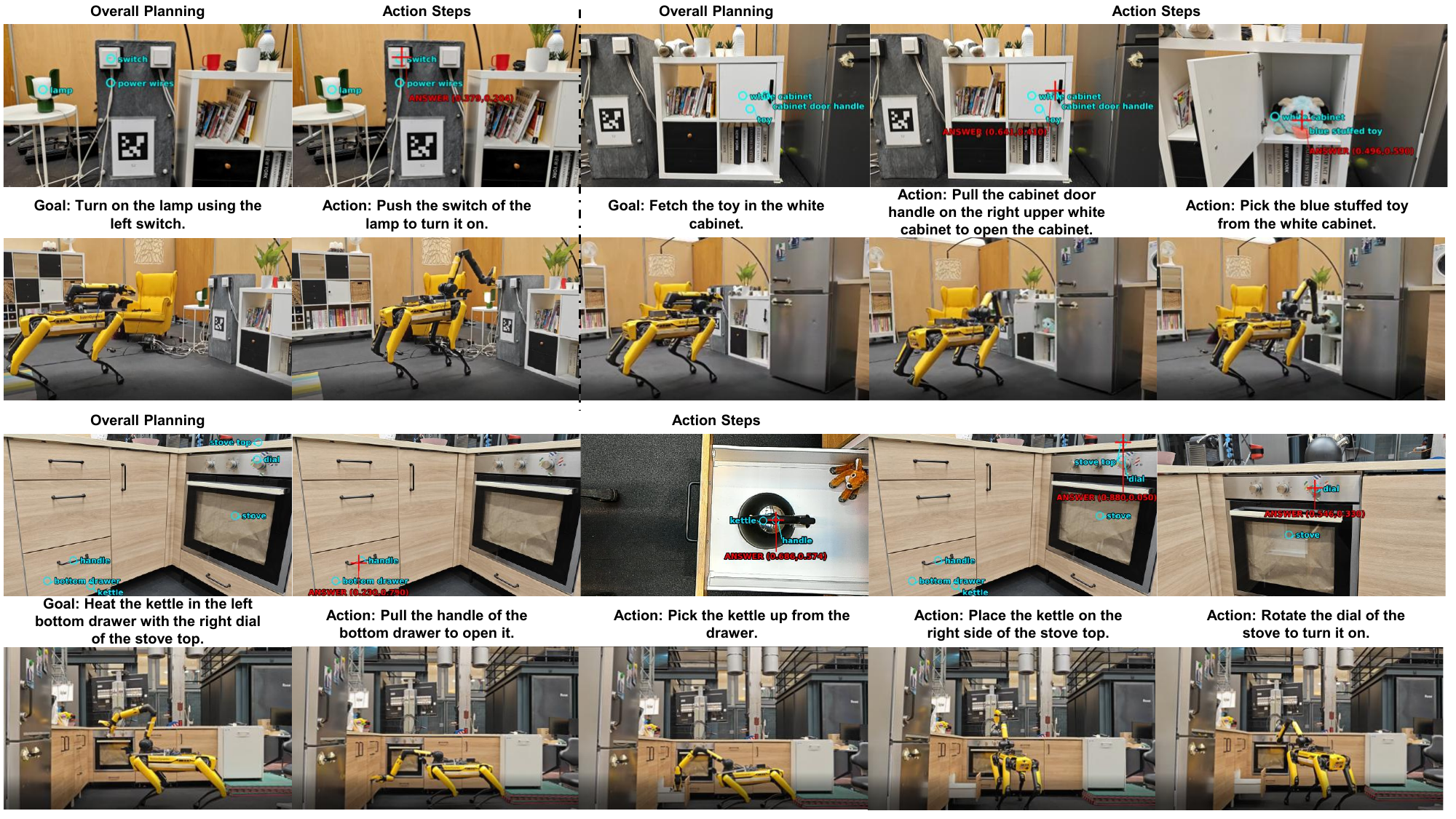}
    % \vspace{-8mm}
    \caption{\textbf{Zero-shot real-world mobile manipulation with a Boston Dynamics Spot robot.}
    ChronoGraphVLM predicts sub-actions and updates affordance grounding before each interaction, guiding tasks of increasing horizon through existing robot skills.}
    \label{fig:robot}
    % \vspace{-5mm}
\end{figure*}

Figure~\ref{fig:robot} shows three tasks of increasing horizon: pushing a switch to activate a lamp; opening a cabinet and grasping the newly exposed toy; and opening a bottom drawer, retrieving a kettle, placing it on the appropriate stovetop region, and rotating the corresponding dial.
These qualitative demonstrations illustrate how goal decomposition and observation-conditioned affordance grounding support sequential manipulation through existing robot skills in an unseen environment.
\section{Conclusion and Limitations}

We introduce ChronoGraph, a functional 4D scene graph that links affordance-level actions to semantic and geometric changes, connecting 4D interaction understanding with spatially grounded planning.
Building on it, we construct ChronoGraphBench from human and robot interactions and train ChronoGraphVLM with Graph-as-Chain-of-Thought supervised fine-tuning and joint 4D graph reinforcement learning.
The resulting models improve over pretrained baselines across scales, transfer zero-shot to VLM4D, and support real-world mobile manipulation through existing robot skills without additional fine-tuning.
These results highlight the value of learning how interactions change the scene and inform subsequent actions.
However, dataset scale is limited by the cost of affordance-level 3D annotation, and inference latency hinders real-time closed-loop replanning.
Future work could broaden coverage across environments and robot platforms, and explore distillation and streaming inference for incremental graph and plan updates.

\subsubsection*{Acknowledgments}
We thank the help provided by the colleagues in developing the project:
Zuria Bauer, Zhengyu Fu, and Tim Engelbracht.

\bibliography{iclr2027_conference}

@inproceedings{chen2024spatialvlm,
  title={Spatialvlm: Endowing vision-language models with spatial reasoning capabilities},
  author={Chen, Boyuan and Xu, Zhuo and Kirmani, Sean and Ichter, Brain and Sadigh, Dorsa and Guibas, Leonidas and Xia, Fei},
  booktitle={Proceedings of the IEEE/CVF Conference on Computer Vision and Pattern Recognition},
  pages={14455--14465},
  year={2024}
}

@article{chen2023egoplan,
  title   = {EgoPlan-Bench: Benchmarking Multimodal Large Language Models for Human-Level Planning},
  author  = {Chen, Yi and Ge, Yuying and Ge, Yixiao and Ding, Mingyu and Li, Bohao and Wang, Rui and Xu, Ruifeng and Shan, Ying and Liu, Xihui},
  journal = {arXiv preprint arXiv:2312.06722},
  year    = {2023}
}

@inproceedings{sermanet2024robovqa,
  title     = {RoboVQA: Multimodal Long-Horizon Reasoning for Robotics},
  author    = {Sermanet, Pierre and Ding, Tianli and Zhao, Jeffrey and Xia, Fei and Dwibedi, Debidatta and Gopalakrishnan, Keerthana and Chan, Christine and Dulac-Arnold, Gabriel and Maddineni, Sharath and Joshi, Nikhil J and Florence, Pete and Han, Wei and Baruch, Robert and Lu, Yao and Mirchandani, Suvir and Xu, Peng and Sanketi, Pannag and Hausman, Karol and Shafran, Izhak and Ichter, Brian and Cao, Yuan},
  booktitle = {IEEE International Conference on Robotics and Automation (ICRA)},
  year      = {2024}
}

@inproceedings{delitzas2024scenefun3d, 
  title = {{SceneFun3D: Fine-Grained Functionality and Affordance Understanding in 3D Scenes}}, 
  author = {Delitzas, Alexandros and Takmaz, Ayca and Tombari, Federico and Sumner, Robert and Pollefeys, Marc and Engelmann, Francis}, 
  booktitle = {IEEE/CVF Conference on Computer Vision and Pattern Recognition (CVPR)}, 
  year = {2024}
}

@article{lin2025depth,
  title={Depth anything 3: Recovering the visual space from any views},
  author={Lin, Haotong and Chen, Sili and Liew, Junhao and Chen, Donny Y and Li, Zhenyu and Shi, Guang and Feng, Jiashi and Kang, Bingyi},
  journal={arXiv preprint arXiv:2511.10647},
  year={2025}
}

@article{cheng2024spatialrgpt,
  title={Spatialrgpt: Grounded spatial reasoning in vision-language models},
  author={Cheng, An-Chieh and Yin, Hongxu and Fu, Yang and Guo, Qiushan and Yang, Ruihan and Kautz, Jan and Wang, Xiaolong and Liu, Sifei},
  journal={Advances in Neural Information Processing Systems},
  volume={37},
  pages={135062--135093},
  year={2024}
}

@inproceedings{zhou2026learning,
  title={Learning to reason in 4d: Dynamic spatial understanding for vision language models},
  author={Zhou, Shengchao and Chen, Yuxin and Ge, Yuying and Huang, Wei and Lin, Jiehong and Shan, Ying and Qi, Xiaojuan},
  booktitle={Proceedings of the IEEE/CVF Conference on Computer Vision and Pattern Recognition},
  pages={9637--9646},
  year={2026}
}

@inproceedings{huang2026thinking,
  title={Thinking in dynamics: How multimodal large language models perceive, track, and reason dynamics in physical 4d world},
  author={Huang, Yuzhi and Wen, Kairun and Gao, Rongxin and Liu, Dongxuan and Lou, Yibin and Wu, Jie and Xu, Jing and Zhang, Jian and Yang, Zheng and Lin, Yunlong and others},
  booktitle={Proceedings of the IEEE/CVF Conference on Computer Vision and Pattern Recognition},
  pages={33446--33456},
  year={2026}
}

@article{ju2025momagraph,
  title={MomaGraph: State-Aware Unified Scene Graphs with Vision-Language Model for Embodied Task Planning},
  author={Ju, Yuanchen and Liang, Yongyuan and Wang, Yen-Jen and Gireesh, Nandiraju and Ju, Yuanliang and Lee, Seungjae and Gu, Qiao and Hsieh, Elvis and Huang, Furong and Sreenath, Koushil},
  journal={arXiv preprint arXiv:2512.16909},
  year={2025}
}

@inproceedings{zhang2025open,
  title={Open-vocabulary functional 3d scene graphs for real-world indoor spaces},
  author={Zhang, Chenyangguang and Delitzas, Alexandros and Wang, Fangjinhua and Zhang, Ruida and Ji, Xiangyang and Pollefeys, Marc and Engelmann, Francis},
  booktitle={Proceedings of the Computer Vision and Pattern Recognition Conference},
  pages={19401--19413},
  year={2025}
}

@article{hu2026hierarchical,
  title={Hierarchical and Holistic Open-Vocabulary Functional 3D Scene Graphs for Indoor Spaces},
  author={Hu, Xinggang and Zhang, Chenyangguang and Delitzas, Alexandros and Zhang, Xiangkui and Pollefeys, Marc and Engelmann, Francis and Ji, Xiangyang},
  journal={arXiv preprint arXiv:2605.15753},
  year={2026}
}

@inproceedings{fu2026funfact,
  title={Funfact: building probabilistic functional 3d scene graphs via factor-graph reasoning},
  author={Fu, Zhengyu and Zurbr{\"u}gg, Ren{\'e} and Qu, Kaixian and Pollefeys, Marc and Hutter, Marco and Blum, Hermann and Bauer, Zuria},
  booktitle={Proceedings of the IEEE/CVF Conference on Computer Vision and Pattern Recognition},
  pages={23848--23858},
  year={2026}
}

@inproceedings{li2024muep,
  title={MuEP: A Multimodal Benchmark for Embodied Planning with Foundation Models.},
  author={Li, Kanxue and Yu, Baosheng and Zheng, Qi and Zhan, Yibing and Zhang, Yuhui and Zhang, Tianle and Yang, Yijun and Chen, Yue and Sun, Lei and Cao, Qiong and others},
  booktitle={IJCAI},
  pages={129--138},
  year={2024}
}

@article{rana2023sayplan,
  title={Sayplan: Grounding large language models using 3d scene graphs for scalable robot task planning},
  author={Rana, Krishan and Haviland, Jesse and Garg, Sourav and Abou-Chakra, Jad and Reid, Ian and Suenderhauf, Niko},
  journal={arXiv preprint arXiv:2307.06135},
  year={2023}
}

@inproceedings{perrett2025hd,
  title={Hd-epic: A highly-detailed egocentric video dataset},
  author={Perrett, Toby and Darkhalil, Ahmad and Sinha, Saptarshi and Emara, Omar and Pollard, Sam and Parida, Kranti Kumar and Liu, Kaiting and Gatti, Prajwal and Bansal, Siddhant and Flanagan, Kevin and others},
  booktitle={Proceedings of the Computer Vision and Pattern Recognition Conference},
  pages={23901--23913},
  year={2025}
}

@article{engelbracht2025hoi,
  title={Hoi!-A Multimodal Dataset for Force-Grounded, Cross-View Articulated Manipulation},
  author={Engelbracht, Tim and Zurbr{\"u}gg, Ren{\'e} and Wohlrapp, Matteo and B{\"u}chner, Martin and Valada, Abhinav and Pollefeys, Marc and Blum, Hermann and Bauer, Zuria},
  journal={arXiv preprint arXiv:2512.04884},
  year={2025}
}

@article{nasiriany2026robocasa365,
  title={Robocasa365: A large-scale simulation framework for training and benchmarking generalist robots},
  author={Nasiriany, Soroush and Nasiriany, Sepehr and Maddukuri, Abhiram and Zhu, Yuke},
  journal={arXiv preprint arXiv:2603.04356},
  year={2026}
}

@article{yin2026mllm,
  title={Mllm-4d: Towards visual-based spatial-temporal intelligence},
  author={Yin, Xingyilang and Li, Chengzhengxu and Chang, Jiahao and Pun, Chi-Man and Cun, Xiaodong},
  journal={arXiv preprint arXiv:2603.00515},
  year={2026}
}

@article{zhou2025robotracer,
  title={Robotracer: Mastering spatial trace with reasoning in vision-language models for robotics},
  author={Zhou, Enshen and Chi, Cheng and Li, Yibo and An, Jingkun and Zhang, Jiayuan and Rong, Shanyu and Han, Yi and Ji, Yuheng and Liu, Mengzhen and Wang, Pengwei and others},
  journal={arXiv preprint arXiv:2512.13660},
  year={2025}
}

@article{wu2026spatial,
  title={Spatial-mllm: Boosting mllm capabilities in visual-based spatial intelligence},
  author={Wu, Diankun and Liu, Fangfu and Hung, Yi-Hsin and Duan, Yueqi},
  journal={Advances in neural information processing systems},
  volume={38},
  pages={13569--13597},
  year={2026}
}

@inproceedings{yang2025thinking,
  title={Thinking in space: How multimodal large language models see, remember, and recall spaces},
  author={Yang, Jihan and Yang, Shusheng and Gupta, Anjali W and Han, Rilyn and Fei-Fei, Li and Xie, Saining},
  booktitle={2025 IEEE/CVF Conference on Computer Vision and Pattern Recognition (CVPR)},
  pages={10632--10643},
  year={2025},
  organization={IEEE}
}

@article{zhang2026embodied3dbench,
  title={Embodied3DBench: Benchmarking Low-Level Embodied Spatial Intelligence of Vision Language Models},
  author={Zhang, Jiyao and Zhang, Mingxu and Peng, Yitong and Liu, Haoxuan and Wang, Chenshuo and Long, Yuxing and Huang, Haoyang and Li, Dongjiang and Duan, Nan and Shen, Hui and others},
  journal={arXiv preprint arXiv:2605.29074},
  year={2026}
}

@article{ma2026spatialreasoner,
  title={Spatialreasoner: Towards explicit and generalizable 3d spatial reasoning},
  author={Ma, Wufei and Chou, Yu-Cheng and Liu, Qihao and Wang, Xingrui and de Melo, Celso and Xie, Jianwen and Yuille, Alan},
  journal={Advances in Neural Information Processing Systems},
  volume={38},
  pages={140751--140774},
  year={2026}
}

@article{shen2026fine,
  title={Fine-grained preference optimization improves spatial reasoning in vlms},
  author={Shen, Yifan and Liu, Yuanzhe and Zhu, Jingyuan and Cao, Xu and Zhang, Xiaofeng and He, Yixiao and Ye, Wenming and Rehg, James and Lourentzou, Ismini},
  journal={Advances in Neural Information Processing Systems},
  volume={38},
  pages={17929--17960},
  year={2026}
}

@inproceedings{xu2026multi,
  title={Multi-spatialmllm: Multi-frame spatial understanding with multi-modal large language models},
  author={Xu, Runsen and Wang, Weiyao and Tang, Hao and Chen, Xingyu and Wang, Xiaodong and Chu, Fu-Jen and Feiszli, Matt and Liang, Kevin J},
  booktitle={Proceedings of the IEEE/CVF Conference on Computer Vision and Pattern Recognition},
  pages={31078--31088},
  year={2026}
}

@article{ouyang2025spacer,
  title={Spacer: Reinforcing mllms in video spatial reasoning},
  author={Ouyang, Kun and Liu, Yuanxin and Wu, Haoning and Liu, Yi and Zhou, Hao and Zhou, Jie and Meng, Fandong and Sun, Xu},
  journal={arXiv preprint arXiv:2504.01805},
  year={2025}
}

@article{huang2024chat,
  title={Chat-scene: Bridging 3d scene and large language models with object identifiers},
  author={Huang, Haifeng and Chen, Yilun and Wang, Zehan and Huang, Rongjie and Xu, Runsen and Wang, Tai and Liu, Luping and Cheng, Xize and Zhao, Yang and Pang, Jiangmiao and others},
  journal={Advances in Neural Information Processing Systems},
  volume={37},
  pages={113991--114017},
  year={2024}
}

@inproceedings{deng20253d,
  title={3d-llava: Towards generalist 3d lmms with omni superpoint transformer},
  author={Deng, Jiajun and He, Tianyu and Jiang, Li and Wang, Tianyu and Dayoub, Feras and Reid, Ian},
  booktitle={2025 IEEE/CVF Conference on Computer Vision and Pattern Recognition (CVPR)},
  pages={3772--3782},
  year={2025},
  organization={IEEE}
}

@inproceedings{zhu2025llava,
  title={Llava-3d: A simple yet effective pathway to empowering lmms with 3d capabilities},
  author={Zhu, Chenming and Wang, Tai and Zhang, Wenwei and Pang, Jiangmiao and Liu, Xihui},
  booktitle={2025 IEEE/CVF International Conference on Computer Vision (ICCV)},
  pages={4295--4305},
  year={2025},
  organization={IEEE}
}

@article{liu2026ssr,
  title={Ssr: Enhancing depth perception in vision-language models via rationale-guided spatial reasoning},
  author={Liu, Yang and Ma, Ming and Yu, Xiaomin and Ding, Pengxiang and Zhao, Han and Sun, Mingyang and Huang, Siteng and Wang, Donglin},
  journal={Advances in Neural Information Processing Systems},
  volume={38},
  pages={123926--123958},
  year={2026}
}

@article{zheng2026learning,
  title={Learning from videos for 3d world: Enhancing mllms with 3d vision geometry priors},
  author={Zheng, Duo and Li, Yanyang and Wang, Liwei and others},
  journal={Advances in neural information processing systems},
  volume={38},
  pages={20560--20586},
  year={2026}
}

@inproceedings{azuma2022scanqa,
  title={Scanqa: 3d question answering for spatial scene understanding},
  author={Azuma, Daichi and Miyanishi, Taiki and Kurita, Shuhei and Kawanabe, Motoaki},
  booktitle={2022 IEEE/CVF Conference on Computer Vision and Pattern Recognition (CVPR)},
  pages={19107--19117},
  year={2022},
  organization={IEEE}
}

@article{ma2022sqa3d,
  title={Sqa3d: Situated question answering in 3d scenes},
  author={Ma, Xiaojian and Yong, Silong and Zheng, Zilong and Li, Qing and Liang, Yitao and Zhu, Song-Chun and Huang, Siyuan},
  journal={arXiv preprint arXiv:2210.07474},
  year={2022}
}

@article{zhang2026flatland,
  title={From flatland to space: Teaching vision-language models to perceive and reason in 3d},
  author={Zhang, Jiahui and Chen, Yurui and Xu, Yueming and Huang, Ze and Mei, Jilin and Chen, Chunhui and Zhou, Yanpeng and Yuan, Yu-Jie and Cai, Xinyue and Huang, Guowei and others},
  journal={Advances in Neural Information Processing Systems},
  volume={38},
  year={2026}
}

@inproceedings{yang2026mmsi,
  title={Mmsi-bench: A benchmark for multi-image spatial intelligence},
  author={Yang, Sihan and Xu, Runsen and Xie, Yiman and Yang, Sizhe and Li, Mo and Lin, Jingli and Zhu, Chenming and Chen, Xiaochen and Duan, Haodong and Yue, Xiangyu and others},
  booktitle={International Conference on Learning Representations},
  volume={2026},
  pages={157051--157088},
  year={2026}
}

@inproceedings{jia2026omnispatial,
  title={Omnispatial: Towards comprehensive spatial reasoning benchmark for vision language models},
  author={Jia, Mengdi and Qi, Zekun and Zhang, Shaochen and Zhang, Wenyao and Yu, Xinqiang and He, Jiawei and Wang, He and Yi, Li},
  booktitle={International Conference on Learning Representations},
  volume={2026},
  pages={35634--35670},
  year={2026}
}

@inproceedings{zhou2025vlm4d,
  title={Vlm4d: Towards spatiotemporal awareness in vision language models},
  author={Zhou, Shijie and Vilesov, Alexander and He, Xuehai and Wan, Ziyu and Zhang, Shuwang and Nagachandra, Aditya and Chang, Di and Chen, Dongdong and Wang, Xin Eric and Kadambi, Achuta},
  booktitle={2025 IEEE/CVF International Conference on Computer Vision (ICCV)},
  pages={8600--8612},
  year={2025},
  organization={IEEE}
}

@inproceedings{li2025sti,
  title={Sti-bench: Are mllms ready for precise spatial-temporal world understanding?},
  author={Li, Yun and Zhang, Yiming and Lin, Tao and Liu, XiangRui and Cai, Wenxiao and Liu, Zheng and Zhao, Bo},
  booktitle={2025 IEEE/CVF International Conference on Computer Vision (ICCV)},
  pages={5622--5632},
  year={2025},
  organization={IEEE}
}

@article{guo2025beyond,
  title={Beyond flatlands: Unlocking spatial intelligence by decoupling 3d reasoning from numerical regression},
  author={Guo, Zhongbin and Liu, Jiahe and Li, Yushan and Gao, Wenyu and Yang, Zhen and Li, Chenzhi and Zhang, Xinyue and Jian, Ping},
  journal={arXiv preprint arXiv:2511.11239},
  year={2025}
}

@article{niu2024llarva,
  title={Llarva: Vision-action instruction tuning enhances robot learning},
  author={Niu, Dantong and Sharma, Yuvan and Biamby, Giscard and Quenum, Jerome and Bai, Yutong and Shi, Baifeng and Darrell, Trevor and Herzig, Roei},
  journal={arXiv preprint arXiv:2406.11815},
  year={2024}
}

@inproceedings{guo2024doremi,
  title={Doremi: Grounding language model by detecting and recovering from plan-execution misalignment},
  author={Guo, Yanjiang and Wang, Yen-Jen and Zha, Lihan and Chen, Jianyu},
  booktitle={2024 IEEE/RSJ International Conference on Intelligent Robots and Systems (IROS)},
  pages={12124--12131},
  year={2024},
  organization={IEEE}
}

@article{huang2023voxposer,
  title={Voxposer: Composable 3d value maps for robotic manipulation with language models},
  author={Huang, Wenlong and Wang, Chen and Zhang, Ruohan and Li, Yunzhu and Wu, Jiajun and Fei-Fei, Li},
  journal={arXiv preprint arXiv:2307.05973},
  year={2023}
}

@article{huang2024rekep,
  title={Rekep: Spatio-temporal reasoning of relational keypoint constraints for robotic manipulation},
  author={Huang, Wenlong and Wang, Chen and Li, Yunzhu and Zhang, Ruohan and Fei-Fei, Li},
  journal={arXiv preprint arXiv:2409.01652},
  year={2024}
}

@inproceedings{zheng2025tracevla,
  title={Tracevla: Visual trace prompting enhances spatial-temporal awareness for generalist robotic policies},
  author={Zheng, Ruijie and Liang, Yongyuan and Huang, Shuaiyi and Gao, Jianfeng and Daum{\'e} III, Hal and Kolobov, Andrey and Huang, Furong and Yang, Jianwei},
  booktitle={International Conference on Learning Representations},
  volume={2025},
  pages={54277--54296},
  year={2025}
}

@inproceedings{dai2024optimal,
  title={Optimal scene graph planning with large language model guidance},
  author={Dai, Zhirui and Asgharivaskasi, Arash and Duong, Thai and Lin, Shusen and Tzes, Maria-Elizabeth and Pappas, George and Atanasov, Nikolay},
  booktitle={2024 IEEE International Conference on Robotics and Automation (ICRA)},
  pages={14062--14069},
  year={2024},
  organization={IEEE}
}

@article{ekpo2024verigraph,
  title={Verigraph: Scene graphs for execution verifiable robot planning},
  author={Ekpo, Daniel and Levy, Mara and Suri, Saksham and Huynh, Chuong and Swaminathan, Archana and Shrivastava, Abhinav},
  journal={arXiv preprint arXiv:2411.10446},
  year={2024}
}

@article{needleman1970general,
  title={A general method applicable to the search for similarities in the amino acid sequence of two proteins},
  author={Needleman, Saul B and Wunsch, Christian D},
  journal={Journal of molecular biology},
  volume={48},
  number={3},
  pages={443--453},
  year={1970},
  publisher={Elsevier}
}

@inproceedings{carion2025sam3segmentconcepts,
  title={Sam 3: Segment anything with concepts},
  author={Carion, Nicolas and Gustafson, Laura and Hu, Yuan-Ting and Debnath, Shoubhik and Hu, Ronghang and Suris Coll-Vinent, Didac and Ryali, Chaitanya and Alwala, Kalyan Vasudev and Khedr, Haitham and Huang, Andrew and others},
  booktitle={International conference on learning representations},
  volume={2026},
  pages={138846--138923},
  year={2026}
}

@inproceedings{sam3dteam2025sam3d3dfyimages,
  title={Sam 3d: 3dfy anything in images},
  author={Chen, Xingyu and Chu, Fu-Jen and Gleize, Pierre and Liang, Kevin J and Sax, Alexander and Tang, Hao and Wang, Weiyao and Guo, Michelle and Hardin, Thibaut and Li, Xiang and others},
  booktitle={Proceedings of the IEEE/CVF Conference on Computer Vision and Pattern Recognition},
  pages={7220--7232},
  year={2026}
}

@article{shao2024deepseekmath,
  title={Deepseekmath: Pushing the limits of mathematical reasoning in open language models},
  author={Shao, Zhihong and Wang, Peiyi and Zhu, Qihao and Xu, Runxin and Song, Junxiao and Bi, Xiao and Zhang, Haowei and Zhang, Mingchuan and Li, YK and Wu, Yang and others},
  journal={arXiv preprint arXiv:2402.03300},
  year={2024}
}

@article{chu2025sft,
  title={Sft memorizes, rl generalizes: A comparative study of foundation model post-training},
  author={Chu, Tianzhe and Zhai, Yuexiang and Yang, Jihan and Tong, Shengbang and Xie, Saining and Schuurmans, Dale and Le, Quoc V and Levine, Sergey and Ma, Yi},
  journal={arXiv preprint arXiv:2501.17161},
  year={2025}
}
\bibliographystyle{iclr2027_conference}

% \clearpage
% \section*{Appendix}
% \input{Sections/8app}

\end{document}